\documentclass[a4paper]{article}

\usepackage[ruled,vlined]{algorithm2e}

\usepackage{INTERSPEECH2019}

\title{Active Annotation: bootstrapping annotation lexicon and guidelines for supervised NLU learning}
\name{Federico Marinelli$^1$$^2$, Alessandra Cervone$^1$, Giuliano Tortoreto$^2$, Evgeny A. Stepanov$^2$,\newline Giuseppe Di Fabbrizio$^2$, Giuseppe Riccardi$^1$}

\address{
  $^1$Signals and Interactive Systems Lab, University of Trento\\
  $^2$VUI Inc.}
  
\email{fdm@vui.com, alessandra.cervone@unitn.it, gtr@vui.com, eas@vui.com, pdf@vui.com, giuseppe.riccardi@unitn.it}

\begin{document}

\maketitle
\begin{abstract}
Natural Language Understanding (NLU) models are typically trained in a supervised learning framework. In the case of intent classification, the predicted labels are predefined and based on the designed annotation schema while the labeling process is based on a laborious task where annotators manually inspect each utterance and assign the corresponding label. We propose an Active Annotation (AA) 
approach where we combine an unsupervised learning method in the embedding space, a human-in-the-loop verification process, and linguistic insights to create lexicons that can be open categories and adapted over time. In particular, annotators define the y-label space on-the-fly during the annotation using an iterative process and without the need for prior knowledge about the input data. We evaluate the proposed annotation paradigm in a real use-case NLU scenario. Results show that our Active Annotation paradigm achieves accurate and higher quality training data, with an annotation speed of an order of magnitude higher with respect to the traditional human-only driven baseline annotation methodology.

\end{abstract}

\section{Introduction}

Supervised methods are arguably one of the most popular and used techniques for a wide range of tasks in Machine Learning (ML), especially Natural Language tasks.
This class of techniques is based on the notion of ``programming by example'', where the annotation guidelines let the designers specify the output that they are looking for. Even unsupervised learning methods require to be evaluated on labelled examples in order to assess their quality.
For a typical natural language annotation task, a language expert would define the annotation guidelines document that usually includes a data schema, annotation instructions and examples. The annotation schema specifies the \textit{y}-label space the ML model has to predict. Instructions and examples instruct annotators about the
rules to follow in order to assign correct text classification labels.
In general, it is difficult to have a clear a priori view of the data that has to be annotated:  having thousand of documents to annotate requires a lot of human effort in order to provide well-defined annotation guidelines. 

In this paper we focus on the following research question: \textit{What is the optimal balance between accuracy and speed in human annotation tasks?}
In particular, we investigate the hypothesis that incorporating the notion of iteration in the annotation process might lead to faster and more accurate annotations, given the difficulty in correctly defining the annotation task a priori. 
To address this challenge, we first introduce the notion of an iterative annotation process where annotations are refined interactively and dynamically. As second step, we focus on semi-automatic annotation \cite{Raymond2008ActiveAI} where machines leverage consistency, memory and recall while humans deal with context, ambiguity and precision.
Additionally, in order to explore the optimal ratio between annotation speed and accuracy, we choose to break down complex annotation tasks into binary questions, that we assume to be simpler and faster to annotate.
This approach allows annotators to speed up the annotation process and, at the same time, measure the reliability of the process.



The paper is structured as follows: we first provide a description of related work (Section 2) in active annotation, while in Section 3 we present a high-level overview of the proposed algorithm. Next, in Section 4 we explain the guidelines definition procedure and describe our proposed annotation process. Finally, we present experimental results (Section 5) and draw on the conclusions of our work (Section 6).

\section{Background}
In this section,  we first discuss relevant work from the emerging field of human-in-the-loop  computational  architectures  applied  for  data  annotation,  then  we  review  existing work methodologically related to our proposed framework.

Human-in-the-loop systems, or human-machine hybrid systems, are aimed at exploiting the complementarity between the intelligence of humans and the scalability of machines to solve complex tasks at scale \cite{DEMARTINI20155}. The number of human-in-the-loop systems proposed recently increased, demonstrating the power of human intelligence when coupled with machines in solving complex tasks for intelligent machines (e.g.  Recaptcha for OCR application \cite{recaptha}). Crowd-sourcing is often used to collect and annotate data to train supervised machine learning models in many natural language processing tasks, such as sentiment  and  opinion  mining  \cite{hitl-2}  and question answering \cite{hithq}.  A very popular and well investigated framework in order to cope with the lack of training material, that uses a human-in-the-loop paradigm, is the active learning paradigm \cite{Cohn1994ActiveLW}.  It has been applied to various NLP tasks with impressive results in  reducing  the  amount of  annotated  training  data. 

Among the applications of Active Learning, the procedure has been combined with annotation error detection for speeding up annotation processes and minimizing the human effort \cite{Raymond2008ActiveAI}. The created procedure, called Active Annotation, improved convergence time to reliable automatic annotation, selecting for annotation the most informative examples, thus reducing the number of training examples needed to achieve a given level of performance through the Active Learning paradigm.
As in Active Annotation, the proposed procedure selects at each turn the most informative examples to be annotated by the user. Contrary to the previous approaches that required a predefined \textit{y}-label space before the annotation process, in our annotation paradigm the labels are created on-the-fly during the annotation, without the need for prior knowledge of the input data. Such paradigm is particularly effective when no labelled data are available.

\section{Active Annotation}
The Active Annotation methodology employs both human and machine intelligence to create data for machine learning models. The final objective is to shorten the time to obtain effective machine learning models from scratch. 
Algorithm 1 shows a general overview of the Active Annotation methodology that we are going to present.

\begin{algorithm}
\SetAlgoLined
\textbf{Input} = \textit{D} -- set of unlabelled examples\;
E = Embeddings-Computation(D)\;
E' = Embeddings-Dimensionality-Reduction(E)\;
$C_{E'}$ = Embeddings-Clustering(E', K)\;
 \While{D not Empty}
 {
  \If{Guidelines-Definition-Procedure($C_{E'}$) == True}
  		{\textit{Annotation-Procedure}(D, $C_{E'}$)}
  
 }
 \caption{Active Annotation Algorithm Overview}
\end{algorithm}

\noindent
\textit{D} -- represents the unlabelled examples that we want to label.

\smallbreak
\noindent
\textit{Embeddings-Computation} --  refers to a function that takes as input the unlabelled examples \textit{D} and returns the relative vector representation for each data point, namely \textit{E}.

\smallbreak
\noindent
\textit{Embeddings-Dimensionality-Reduction} -- represents a function that takes as input a high dimensional vector representation of the input data, the embedding \textit{E}, and returns a vector with lower dimensions for each data point, namely \textit{E'}.

\smallbreak
\noindent
\textit{Embeddings-Clustering} -- is a function that takes as input the vector representation of the data and clusters it according to the given clustering methodology. The output of this function are \textit{K} clusters of the given data, namely $C_{E'}$.

\smallbreak
\noindent
\textit{Guidelines-Definition Procedure} -- is the process by which the annotator, given \textit{N} key data points and an automatically computed cluster label, is asked to provide a \textit{y}-label to the given \textit{N} data points. The annotators can either provide a label or skip if they do not know how to annotate the proposed data points.

\smallbreak
\noindent
\textit{Annotation-procedure} -- is the process that follows the Guidelines-Definition procedure. In case the annotator provides a \textit{y}-label during the previous step, the algorithm proposes \textit{K} data points, that are very similar to the \textit{N} data points that the annotator has already labelled during the previous step. The annotator is asked to analyze the proposed \textit{K} data points and annotate them through a binary decision process.

\smallbreak
\noindent
The algorithms initially represents the unlabelled data-points in a distributed representation, vectors, so that Machine Learning models can easily process them. 
We used the ``Universal Sentence Encoder'' \cite{Cer2018UniversalSE} algorithm to transform sentences into vectors. On the encoded sentence, we apply a dimensionality reduction algorithm. Therefore the number of parameters provided to the Active Learning component is decreased, as well as the noise and the sparsity of the data, effectively leading to a faster training process. 
Speed is a major concern, since the proposed Active Annotation paradigm is served through web server, keeping system latency low is important to support the annotation process. 
To perform dimensionality reduction we apply the Principal Component Analysis (PCA) \cite{abdi2010principal}, on the resulting data we perform unsupervised clustering using K-Means clustering, in particular the k-means++ \cite{kmeans++} implementation. K-means++ provides faster and better performances compared to the standard algorithm \cite{kmeans++}, by selecting one cluster centroid randomly and then searching for other centroids.
The selection of the number of clusters, $k$, in k-means algorithm is essential for our task, since the created clusters drive the annotators experience.
Selecting the correct $k$ is complex and it is affected by the shape, scale, the distribution of data points and the level of detail required by the user.
To choose the optimal $k$ we apply the ``Elbow Method'' \cite{bholowalia2014ebk} aiming to balance between maximum compression of the data using a single cluster, and maximum accuracy by assigning each data point to its own cluster. After the clustering phase, we start the iterative annotation process, that is the core procedure of the Active Annotation methodology. It is composed of two main phases, that are called the ``\textit{Guidelines-Definition-Procedure}'' and the ``\textit{Annotation-Procedure}'', respectively. We are going to describe in details these two phases in the following section.

\section{Algorithms}
\subsection{Guidelines Definition Procedure}
\begin{algorithm}
\SetAlgoLined
\textbf{Input} = $C_{E'}$ -- Clustered input data\;
Select one random cluster \textit{c} inside $C_{E'}$\;
Select the \textit{N} most informative data points in \textit{c} as Pivots\;
Compute a cluster label $L_c$\;
Propose the \textit{N} Pivots data points and $L_c$ to the annotator\;
\eIf{Annotator provides a label}
		{return True\;}
     	{return False\;}
 \caption{Guidelines Definition Pseudo Algorithm}
\end{algorithm} 

\smallbreak
\noindent
The Guidelines definition procedure is one of the two core components of the Active Annotation paradigm. Given a number of $k$ clusters, the algorithm selects randomly a cluster \textit{c}. Once a cluster \textit{c} has been selected, the algorithm selects the \textit{N} most representative data points within the cluster \textit{c} as \textit{Pivot} data points. \textit{Pivots} are selected choosing the \textit{N} points closest to the cluster \textit{c}'s centroid. The intuition is that since \textit{Pivot} data points are the nearest to the centroid, they also are the most similar among each others, hence most probably they have the same label.  In this case the decision of the hyper-parameter \textit{N} depends highly on how much cognitive effort we want the annotator to expose during the annotation. In our experiments we set \textit{N} to 3. After that, the algorithm also computes automatically a cluster label $L_c$ (Section 4.2). Once a cluster label $L_c$  and the \textit{N Pivot} data points are selected, the procedure presents them to annotators. They can either decide to provide a label, that could be equal or not to the one automatically computed $L_c$, or can decide to skip. In case the annotator decides to go ahead without providing a label, the algorithm proceeds by selecting a new cluster \textit{c} randomly. Then it computes a new cluster label $L_c$ and selects \textit{N} new data points inside the last selected cluster and finally presents them to the annotator again. Given the iterative nature of the guidelines definition phase, we also call it \textit{exploration phase}: the annotator may quickly explore the real distribution of the labels within the input data just by iteratively inspecting \textit{Pivot} data points inside the clusters. This is based on the assumption that it is likely that data points inside the same cluster have the same label, conversely, data points belonging to different clusters have different labels. Moreover, by selecting a cluster randomly at each iteration it is likely that the next selected cluster will have a different label with respect to the last selected cluster, which, in our opinion, will help annotators to provide better annotations.

\subsection{Predicate-Argument Label Extraction Procedure}
\begin{algorithm}
\SetAlgoLined
\textbf{Input} = One cluster \textit{c}  $\in C_{E'}$ (clustered input data)\;
\ForEach{Sentence \textit{S} $\in c$}{
	SVO += Subject-Verb-Object-Triplet(S) 
}
Lemmatization-procedure(SVO) \;
Stop-words-removal-procedure(SVO) \;
\textit{Predicate} = most-common-verb(SVO)\;
\textit{Argument} = most-common-object(SVO)\;
return string(\textit{Predicate\_Argument})\;
 \caption{Predicate-Argument Extraction Algorithm}
\end{algorithm} 
\smallbreak
\noindent
The Algorithm 3 shows the procedure that we developed in order to automatically extract a cluster label inside a given cluster \textit{c}. 
In current work, as already discussed, we use the Active Annotation Methodology in order to generate labelled data for supervised text classification tasks. We tackle the task of discovering user intents in utterances addressed to a conversational interaction system. A user intent describes what users are looking for when they perform a search query or while they are interacting with a conversational agent.  We believe that a good way to represent the intent of a sentence is to use a predicate-argument structure \cite{surdeanu2003using}. In particular,  a predicate identifies a relation between entities denoted by the subject and complements; an argument is an expression that helps complete the meaning of a predicate that refers to the object of the sentence. An example could be the utterance``\textit{I'd like to add those items to the shopping-cart}'' where the argument is ``\textit{shopping-cart}'' and the predicate is represented by the verb ``\textit{add}''. 
In order to extract such predicate-argument intent-label from all the sentences within a cluster \textit{c} the algorithm extracts subject-verb-object (SVO) triplets from all the sentences in \textit{c}. Then stop-words are removed, and verbs and the objects are lemmatized. The extracted SVO triplets are parsed and the most frequent predicate and argument are selected. If the procedure does not find an argument and a predicate among all the sentences it returns the label ``inform\_\textit{none}''.
The off-the-shelf tools are used to extract the SVO triplets.\footnote{https://spacy.io} 

\subsection{Annotation Procedure}
Once annotators have provided a \textit{y}-label for the proposed \textit{N Pivot} data-points during the guidelines-definition procedure the Active Annotation algorithm proceeds with the so called ``Annotation Procedure''. In this phase, the algorithm proposes the \textit{K} nearest data points to the \textit{N} \textit{Pivots} previously selected. The selection of the data point is performed applying K-Nearest-Neighbours Search algorithm \cite{2016arXiv160309320M}.\footnote{https://github.com/nmslib/nmslib}
The number of proposed data point is arbitrarily set to 5, but it can be increased by annotators through UI up to a predefined threshold. During this phase a binary decision process is employed for each proposed data point, in which annotators are asked to check-mark within the annotation tool the proposed sentences that belong to the \textit{y}-label of the \textit{N Pivot} data points. When the annotation of the proposed sentences is complete, they can proceed to the next Active Annotation iteration by clicking a confirm button. When the button is clicked, check-marked sentences are added to the labelled pool and are removed from the unlabelled data points.

\section{Experiments}
\subsection{Active Annotation Web-tool}
We deployed the Active Annotation paradigm as an annotation web-tool where the annotator can annotate the unlabelled data points. 
The annotation web-tool needs to have several requirements in order to enable fast, accurate and reliable annotations. In particular, we designed a well-structured User Interface (UI)
focusing on annotator's needs, accessibility of elements, and simplicity.
We avoid using long text paragraphs, guiding the UI navigation through the graphical disposition of the objects. We also strategically tried to use colors and textures in order to direct the attention of the user during the annotation process. The web-annotation tool also requires a back-end part that constantly performs computations in order to enable the Active Learning paradigm to help the user during the annotation process.  The back-end has to be efficient and fast: it is of primary importance that the annotation process is not slowed down by the back-end computations. The Annotation Web-Tool was developed using a framework called ``Dash Plotly'' \footnote{https://dash.plot.ly/}. 
Dash app code is declarative and reactive, which makes it easy to build complex apps that contain many interactive elements.

\subsection{Baseline}
To assess the effectiveness of the active annotation tool, we have developed a baseline annotation tool. Similarly to the Active Annotation Methodology, in the baseline we pre-compute a predicate-argument label for each sentence in the unlabelled data to annotate using Algorithm 3. 
Thus, only the annotation methodology is compared and not the automatic labelling algorithm. 

In the baseline, randomly selected sentences are presented one at a time together with the pre-computed predicate-argument labels. The design of the interface is identical to the guidelines-definition of the active annotation web tool.  Here the user is asked to read the sentence, the cluster-label $L_c$ and decide either to: confirm the proposed automatically computed label, provide a new label, or skip.
Once the user has confirmed a label for the proposed sentences, the algorithm adds the sentence-label pairs in the labelled pool and removes it from the unlabelled set \textit{D}. 

\subsection{Dataset}
The dataset used for experiments is the movie-ticket booking conversations from human-human ``e2e Dialogue Datasets'' \cite{li2018microsoft}. 
We have selected 2,140 user turns from the dialogues, maintaining the distribution of the labels. We annotated this data with new intent labels, using a predicate-argument \textit{y}-label structure with 14 labels in total. Labels were verified by three domain expert annotators. We then selected as test-set 140 sentences, with 10 sentences for each \textit{y}-label. The remaining 2,000 sentences are used as unlabelled dataset \textit{D}.

\subsection{Experimental Design}
In order to evaluate the performance of the Active Annotation paradigm presented in this work we selected four annotators. The annotators did not have any prior knowledge about the annotation tasks they had to perform. 
We conducted two experiments with each annotator, each one lasting 25 minutes. In each experiment annotators were asked to annotate the unlabelled dataset \textit{D}: in the first experiment we used the baseline annotation web-tool and in the second experiment we used the Active Annotation web-tool. We dedicated 10 minutes to train annotators to use the annotation web tools. Annotators were asked to use the Baseline annotation web-tool, for 25 minutes, in order to label as many sentence in \textit{D} as possible. After that, they were asked to label as many sentences as possible in \textit{D} using the Active Annotation web-tool, even here for 25 minutes. In particular, the first two annotators used first the baseline web-tool and then the active annotation web-tool, while for the other two annotators the order was reversed. 

\subsection{Results}

We compare the results obtained using the Active Annotation paradigm with those obtained with the Baseline annotation paradigm. In Table 1 we report the mean and the standard deviation of how many data points the annotators were able to annotate during each experiment, with the relative number of labels. In order to assess the quality of the labelled data, we trained a CNN classifier \cite{Kim2014ConvolutionalNN} to predict the intent labels and evaluated the resulting model performance by using a stratified 5-fold cross-validation methodology. To mitigate eventual issues with the relatively small training sample, we randomly extracted the same amount of data annotated in the baseline annotation experiments. The evaluation metric is the F1-score. In Table 2 we report the evaluation of the data labelled, during each experiment, over the test-set that we created (discussed in section 5.3). In order to do such evaluation, since during each experiment the annotators end up with different labels, we manually mapped the labels provided by the annotators with the ground-truth labels on the test-set, where possible. On average we were able to manually map the 80\% of the intent-labels. We report the mean and standard-deviation of the inter-annotation agreement, computed using the Cohen's kappa coefficient \cite{mchugh2012interrater}, between the annotations provided by the annotators and the ground truth annotations. For each experiment we also trained a CNN intent classifier \cite{Kim2014ConvolutionalNN} on the labelled data and we evaluated the classifier over the test-set. Also in this case, we manually mapped the labels provided during the experiments with the ground-truth labels on the test-set. Using the Active Annotation paradigm we achieve better F$_1$-score, both in the cross-validation evaluation and in the test-set evaluation. Also the annotation-agreement is better by using the Active Annotation methodology. Moreover, using our technique, the annotators were able to annotate a number of sentences of an order of magnitude higher with respect to the baseline annotation paradigm.

\begin{table}[]
\centering
\begin{tabular}{|c|c|c|c|c|}
\cline{1-5}
                                                          & \multicolumn{2}{|c|}{\textbf{Baseline}} & \multicolumn{2}{|c|}{\textbf{AA}} \\ 
                                                          & \textit{$\mu$}         & \textit{$\sigma$}        & \textit{$\mu$}             & \textit{$\sigma$}             \\ \hline
\multicolumn{1}{|c|}{Sentences labelled}         & 118.6              & 18.5              & 999.3                  & 171.4                  \\ \hline
\multicolumn{1}{|c|}{Number of labels}           & 10.3               & 1.2               & 8.6                    & 0.9                    \\ \hline
\multicolumn{1}{|c|}{Cross-Validation F1}        & 0.83               & 0.05              & 0.91                  & 0.02                  \\ \hline
\end{tabular}
\smallbreak
\smallbreak
\caption{Mean and standard-deviation results of the four experiments. We report the total number of sentences labelled, the total number of labels provided by the annotators and the stratified 5-fold cross-validation F1 computed on the labelled data generated during each experiment. }
\end{table}

\begin{table}[]
\centering
\begin{tabular}{|c|c|c|c|c|}
\cline{1-5}
                                                          & \multicolumn{2}{c|}{\textbf{Baseline}} & \multicolumn{2}{c|}{\textbf{AA}} \\ 
                                                          & \textit{$\mu$}         & \textit{$\sigma$}        & \textit{$\mu$}             & \textit{$\sigma$}             \\ \hline
\multicolumn{1}{|c|}{F1 Test-Set}        & 0.81               & 0.04              & 0.89                   & 0.03                  \\ \hline
\multicolumn{1}{|c|}{Annotation Agreement} & 0.61               & 0.02              & 0.64                   & 0.01                   \\ \hline
\end{tabular}
\smallbreak
\smallbreak
\caption{Mean and standard-deviation results of the four experiments. We report the evaluation of the CNN intent classification model trained with the labelled data generated during each experiment and evaluated on the test-set. We also report the annotation-agreement, computed using the Cohen's kappa coefficient, between the labels provided during each experiment with those on the ground truth annotations. }
\end{table}

\section{Conclusions}
We presented an Active Annotation (AA) paradigm where we combined unsupervised learning in the embedding space, a human-in-the-loop methodology and linguistic insights to create data for machine learning models. This methodology was evaluated in a real use-case Natural Language Understanding scenario: four internal annotators were enrolled in order to annotate a pool of sentences with intents as target variables. The Active Annotation has been compared with respect to traditional human-only driven baseline annotation methodology. The results showed that the quality of the annotations is improved by leveraging the Active Annotation paradigm, yielding both better inter-annotator agreement and allowing the text-classification model trained with the Active Annotation labelled data to reach better F1-score performances on the test-set. Moreover, the annotation speed is faster when leveraging the Active Annotation paradigm: in all experiments annotators were able to annotate a number of sentences of an order of magnitude higher compared to our baseline. Also, we observed a lower number of end labels when using the AA paradigm. All of the above improvements lead us to the conclusion that Active Annotation is an efficient and powerful methodology that can reduce the time required to obtain training data for machine learning models and gives the possibility to create lexicons that can be open classes and adapted over time.

\bibliographystyle{plain} 

\begin{thebibliography}{10}

\bibitem{abdi2010principal}
Herv{\'e} Abdi and Lynne~J Williams.
\newblock Principal component analysis.
\newblock {\em Wiley interdisciplinary reviews: computational statistics},
  2(4):433--459, 2010.

\bibitem{kmeans++}
David Arthur and Sergei Vassilvitskii.
\newblock K-means++: The advantages of careful seeding.
\newblock In {\em Proceedings of the Eighteenth Annual ACM-SIAM Symposium on
  Discrete Algorithms}, SODA '07, 2007.

\bibitem{bholowalia2014ebk}
Purnima Bholowalia and Arvind Kumar.
\newblock Ebk-means: A clustering technique based on elbow method and k-means
  in wsn.
\newblock {\em International Journal of Computer Applications}, 105(9), 2014.

\bibitem{Cer2018UniversalSE}
Daniel Cer, Yinfei Yang, Sheng yi~Kong, Nan Hua, Nicole Limtiaco, Rhomni~St.
  John, Noah Constant, Mario Guajardo-Cespedes, Steve Yuan, Chris Tar,
  Yun-Hsuan Sung, Brian Strope, and Ray Kurzweil.
\newblock Universal sentence encoder.
\newblock {\em CoRR}, abs/1803.11175, 2018.

\bibitem{Cohn1994ActiveLW}
David~A. Cohn, Zoubin Ghahramani, and Michael~I. Jordan.
\newblock Active learning with statistical models.
\newblock {\em J. Artif. Intell. Res.}, 4:129--145, 1994.

\bibitem{DEMARTINI20155}
Gianluca Demartini.
\newblock Hybrid human–machine information systems: Challenges and
  opportunities.
\newblock {\em Computer Networks}, 90:5 -- 13, 2015.
\newblock Crowdsourcing.

\bibitem{hithq}
Michael Heilman and Noah~A. Smith.
\newblock Rating computer-generated questions with mechanical turk.
\newblock In {\em Mturk@HLT-NAACL}, 2010.

\bibitem{Kim2014ConvolutionalNN}
Yoon Kim.
\newblock Convolutional neural networks for sentence classification.
\newblock In {\em EMNLP}, 2014.

\bibitem{li2018microsoft}
Xiujun Li, Sarah Panda, Jingjing Liu, and Jianfeng Gao.
\newblock Microsoft dialogue challenge: Building end-to-end task-completion
  dialogue systems.
\newblock {\em arXiv preprint arXiv:1807.11125}, 2018.

\bibitem{recaptha}
Colin McMillen David~Abraham Luis Von~Ahn, Benjamin~Maurer and Manuel Blum.
\newblock Recaptcha: Human-based character recognition via web security
  measures.
\newblock {\em Science 321, 5895 (2008), 1465–1468}, 2008.

\bibitem{2016arXiv160309320M}
Y.~A. {Malkov} and D.~A. {Yashunin}.
\newblock {Efficient and robust approximate nearest neighbor search using
  Hierarchical Navigable Small World graphs}.
\newblock {\em arXiv e-prints}, March 2016.

\bibitem{mchugh2012interrater}
Mary~L McHugh.
\newblock Interrater reliability: the kappa statistic.
\newblock {\em Biochemia medica: Biochemia medica}, 22(3):276--282, 2012.

\bibitem{hitl-2}
Bart Mellebeek, Francesc Benavent, Jens Grivolla, Joan Codina, Marta
  R.~Costa-Juss{\`a}, and Rafael Banchs.
\newblock Opinion mining of spanish customer comments with non-expert
  annotations on mechanical turk.
\newblock In {\em Proceedings of the NAACL HLT 2010 Workshop on Creating Speech
  and Language Data with Amazon's Mechanical Turk}, pages 114--121. Association
  for Computational Linguistics, 2010.

\bibitem{Raymond2008ActiveAI}
Christian Raymond, Kepa~Joseba Rodr{\'i}guez, and Giuseppe Riccardi.
\newblock Active annotation in the luna italian corpus of spontaneous
  dialogues.
\newblock In {\em LREC}, 2008.

\bibitem{surdeanu2003using}
Mihai Surdeanu, Sanda Harabagiu, John Williams, and Paul Aarseth.
\newblock Using predicate-argument structures for information extraction.
\newblock In {\em Proceedings of the 41st Annual Meeting of the Association for
  Computational Linguistics}, 2003.

\end{thebibliography}

\end{document}